\RequirePackage{fix-cm}
\documentclass[SNCS]{article}                     
\usepackage{graphicx}
\usepackage{mathptmx}      
\usepackage[utf8]{inputenc}
\usepackage{amsmath}
\usepackage{algorithm}
\usepackage{listings}
\usepackage{placeins} 
\usepackage[noend]{algpseudocode}
\usepackage{cite}

\title{Parallelized Training of Restricted Boltzmann Machines using Markov-Chain Monte Carlo Methods}

\author{Pei Yang                \and
        Srinivas Varadharajan   \and
        Lucas A. Wilson         \and
        Don D. Smith II         \and
        John A. Lockman III     \and
        Vineet Gundecha         \and
        Quy Ta 
}

\begin{document}

\maketitle

\begin{abstract}
Restricted Boltzmann Machine (RBM) is a generative stochastic neural network that can be applied to collaborative filtering technique used by recommendation systems. Prediction accuracy of the RBM model is usually better than that of other models for recommendation systems. However, training the RBM model involves Markov-Chain Monte Carlo (MCMC) method, which is computationally expensive. In this paper, we have successfully applied distributed parallel training using Horovod framework to improve the training time of the RBM model. Our tests show that the distributed training approach of the RBM model has a good scaling efficiency. We also show that this approach effectively reduces the training time to little over 12 minutes on 64 CPU nodes compared to 5 hours on a single CPU node. This will make RBM models more practically applicable in recommendation systems.


\end{abstract}

\section{Introduction}
\label{intro}
Restricted Boltzmann machine (RBM) was first invented in 1980s by Smolensky et al \cite{001_Smolensky}, and intensively studied by Hinton et al \cite{002_Deep_Boltzmman_Machine, 005_RBM_Hinton, 006_RBM_Learning_ReLearning}. Theoretically, RBM has one visible layer and several hidden layers (one hidden layer in most cases), with mutual connections between neurons in different layers while connections between neurons within the same layer are prohibited, as is shown in Figure \ref{fig:RBM architecture original}. Connections between neurons are determined in a way such that the "energy" for the system consisting of all these neurons is minimal. So RBM is an energy-based bidirectional graphical model, whose principles and topologies are quite different from those of other neural networks such as multilayer perceptron (MLP), convolutional neural network (CNN), recurrent neural network (RNN), et al. 

One of the popular applications of RBM is collaborative filtering for recommendation system \cite{003_RBM_for_CF}, where the algorithm needs to predict users' interest levels for products that they have not purchased based on the observed ratings for other products. RBM model outperforms other models for collaborative filtering (e.g., singular value decomposition (SVD) model \cite{004_SVD_for_CF}) by predicting with better accuracy \cite{003_RBM_for_CF}. Considering large datasets with number of users and products(typically more than $100,000$), the number of ratings involved is at the scale of $10^{12}$ or even bigger, which also requires a large memory space to train RBM models. Also, the training algorithm involves Markov Chain Monte Carlo (MCMC) step, which is very computationally expensive. Hence, distributed training is a necessity in order to speed up the training process and practically leverage RBM models for recommendation problems in e-commerce, retails, online entertainment, et al.

One of the efficient algorithms to train the RBM is the contrastive divergence (CD) algorithm initially proposed by Hinton et al \cite{CarreiraPerpin2005OnCD}. The basic idea behind CD algorithm is to approximately draw samples from a joint distribution via sampling from a Markov chain with up to a limited number of steps. CD algorithm has been proved to work well even with just a few steps of Markov chain \cite{003_RBM_for_CF}. However, for a large scale training data, it still takes a long time for the RBM model to converge to a good solution. Hence, it is necessary to explore the parallel training techniques in-order to reduce the time-to-train a model. This would make RBMs to be practically applied for collaborative filtering in recommendation systems. This would also be useful in recommendation systems where there is a need to retrain the model frequently and constrained by the available time. In this paper, for the first time we've succesfully applied parallelized training approach using Horovod \cite{sergeev2018horovod} framework to significantly scale-up RBM models with large scale data sets for collaborative filtering in recommendation systems.

The paper is organized as follows. In Section \ref{RBM architecture}, the model architecture and mathematical principles of RBM are presented. Section \ref{learning algorithm} details the learning algorithm for RBM. Sections \ref{distributed training} and \ref{inference} describe how to perform parallelized training of RBM model and how to make predictions with a trained RBM model. In Section \ref{experiments}, some experimental results with the MovieLens data set \cite{MovieLens} are presented. In the end, Section \ref{conclusion} presents our conclusions.

\section{Restricted Boltzmman Machine for Collaborative Filtering}
\label{RBM architecture}
\begin{figure}[h]
\centering
\includegraphics[scale=0.5]{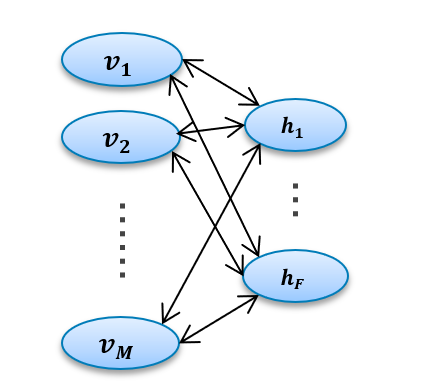}
\caption{RBM architecture}
\label{fig:RBM architecture original}
\end{figure}

\begin{figure}[h]
\centering
\includegraphics[scale=0.35]{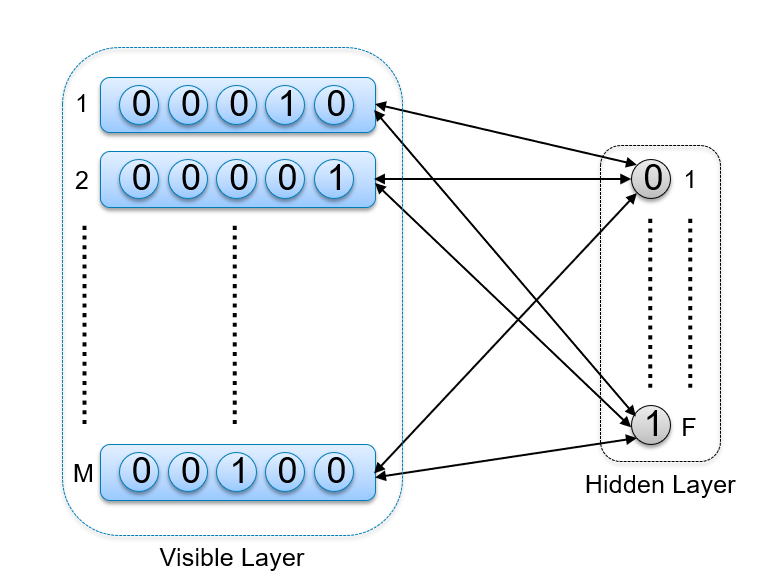}
\caption{RBM architecture (5-way softmax in visible layer)}
\label{fig:RBM architecture 5-way softmax}
\end{figure}

Usually a RBM is a bidirectional network with one visible layer and one hidden layer. The neurons in visible and hidden layers are mutually connected, while connections between neurons within the same layers are restricted, as is shown in Figure \ref{fig:RBM architecture original}. If we try to predict the users' ratings for some products using the collaborative filtering, the visible layer represents the ratings in a 5-way 0's and 1's as shown in Figure \ref{fig:RBM architecture 5-way softmax}.

Suppose we have $N$ users and $M$ products. The $N$ users have rated a portion of the $M$ products, with rating values between $1$ and $K$ ($K=5$ for the case in Figure \ref{fig:RBM architecture 5-way softmax}). For example, if a visible neuron is in the state of $[0, 0, 0, 1, 0]$, it suggests that the user has provided a rating value of $4$ for this product. The visible layer has $M$ neurons, with each corresponding to one of the $M$ products. The states of hidden neurons are binary ($0$ or $1$).

RBM is an energy-based model. For the system $(\boldsymbol{V}, \boldsymbol{H})\ (\boldsymbol{V}=\{v_i^k\}, \boldsymbol{H}=\{h_j\})$, the "energy" is defined as
\begin{align}
    E(\boldsymbol{V}, \boldsymbol{H})=&-\Sigma_{i=1}^m\Sigma_{j=1}^F\Sigma_{k=1}^K v_{i}^k W_{ij}^k h_j \nonumber \\ 
            &- \Sigma_{i=1}^m\Sigma_{k=1}^K v_i^k b_i^k - \Sigma_{j=1}^F h_j c_j
\end{align}
where $W_{ij}^k$ models the interactions between visible and hidden layers, while $b_i^k$ and $c_j$ are bias terms for visible and hidden layers and $F$ denotes the number of neurons in the hidden layer. The joint probability distribution is
\begin{align}
    p(\boldsymbol{V}, \boldsymbol{H})=\frac{exp(-E(\boldsymbol{V}, \boldsymbol{H}))}{Z}
\end{align}
where $Z=\Sigma_{\boldsymbol{V}}\Sigma_{\boldsymbol{H}}p(\boldsymbol{V}, \boldsymbol{H})$ is the normalization factor. It can be shown that \cite{003_RBM_for_CF, 004_Yu}
\begin{equation}
    p(v_i^k=1|\boldsymbol{H}) = \frac{exp(b_i^k + \Sigma_{j=1}^Fh_jW_{ij}^k)}{\Sigma_{l=1}^Kexp(b_i^l + \Sigma_{j=1}^Fh_jW_{ij}^l)}
\end{equation}
and 
\begin{equation}
    p(h_j=1|\boldsymbol{V}) = \sigma(c_j + \Sigma_{i=1}^{m}\Sigma_{k=1}^{K}v_i^kW_{ij}^k)
\end{equation}
where $\sigma(x)=\frac{1}{1+exp(-x)}$ is the sigmoid function. Also, in \cite{004_Yu} it was shown that
\begin{align}
    p(\boldsymbol{V}; \Theta)&=\frac{f({\boldsymbol{V}; \Theta})}{Z}
\end{align}
where $\Theta=(W_{ij}^k, b_i^k, c_j)$ are the parameters for RBM model, and
\begin{align}
    f({\boldsymbol{V}; \Theta}) &= \Sigma_{h_1, h_2, ..., h_F} exp(-E(\boldsymbol{V}, \boldsymbol{H})) \nonumber \\
                                &= \Sigma_{j'=1}^F \Sigma_{h_{j'}=0}^1 exp(\Sigma_{i=1}^m\Sigma_{j=1}^F\Sigma_{k=1}^K v_{i}^k W_{ij}^k h_j \nonumber \\
                                &\ \ \ + \Sigma_{i=1}^m\Sigma_{k=1}^K v_i^k b_i^k + \Sigma_{j=1}^F h_j c_j) \label{f(V, Theta)}
\end{align}
where $\boldsymbol{H}=(h_1, h_2, ..., h_F)$. A detailed description of RBM can be found in \cite{004_Yu}.

\section{Learning Algorithm for RBM}\label{learning algorithm}
Suppose we have observed data $\{\boldsymbol{V}_n\}_{n=1}^N$, then the likelihood with respect to such data is
\begin{align}
    L(\{\boldsymbol{V}_n\}_{n=1}^N, \Theta) &= \Pi_{n=1}^N p(\boldsymbol{V}_n; \Theta) \nonumber \\
                                    &= \Pi_{n=1}^N \frac{f(\boldsymbol{V}_n; \Theta)}{Z(\Theta)}.
\end{align}
To maximize $L(\{\boldsymbol{V}_n\}_{n=1}^N, \Theta)$ is equivalent to minimizing the following objective function
\begin{align}
    G(\{\boldsymbol{V}_n\}_{n=1}^N; \Theta) &= -\frac{1}{N} log(L(\{\boldsymbol{V}_n\}_{n=1}^N; \Theta)) \nonumber \\
     &= log(Z(\Theta)) - \frac{1}{N}\Sigma_{n=1}^N log(f(\boldsymbol{V}_n; \Theta)).
\end{align}

\subsection{Learning via Gradient Descent}
The gradient of $G$ with respect to $\Theta$ is

\begin{equation}
\begin{aligned}
\frac{\partial G}{\partial \Theta} = & \frac{\partial log(Z(\Theta)) }{\partial{\Theta}} - \frac{1}{N}\Sigma_{n=1}^{N}\frac{\partial log(f(\boldsymbol{V}_n; \Theta)) }{\partial{\Theta}} \\
= & \frac{\partial log(Z(\Theta)) }{\partial{\Theta}} - <\frac{\partial log(f(\boldsymbol{V}; \Theta)) }{\partial{\Theta}} >_{\boldsymbol{V}\in \{\boldsymbol{V}_n\}_{i=n}^N}.
\end{aligned}
\end{equation}
The second term is the expectation of $log(f(\boldsymbol{V}; \Theta))$ with respect to observed data $\{\boldsymbol{V}_n\}_{n=1}^N$. For the first term, we have
\begin{align}
\frac{\partial log(Z(\Theta)) }{\partial{\Theta}} &= \frac{1}{Z(\Theta)} \frac{\partial Z(\Theta)}{\partial \Theta} \nonumber \\
                                                &= \frac{1}{Z(\Theta)} \Sigma_{\boldsymbol{V}} \frac{\partial f(\boldsymbol{V}, \Theta)}{\partial \Theta} \nonumber \\
                                                &= \Sigma_{\boldsymbol{V}} \frac{f(\boldsymbol{V}; \Theta)}{Z(\Theta)} \frac{1}{f(\boldsymbol{V}, \Theta)} \frac{\partial f(\boldsymbol{V}, \Theta)}{\partial \Theta} \nonumber \\
                                                &= \Sigma_{\boldsymbol{V}} p(\boldsymbol{V}; \Theta) \frac{\partial logf(\boldsymbol{V}, \Theta)}{\partial \Theta} \nonumber \\
                                                &= <\frac{\partial logf(\boldsymbol{V}; \Theta)}{\partial \Theta}>_{p(\boldsymbol{V}; \Theta)}.
\end{align}
That is, $\frac{\partial log(Z(\Theta)) }{\partial{\Theta}}$ is the expectation of $\frac{\partial logf(\boldsymbol{V}, \Theta)}{\partial \Theta}$ with respect to distribution $p(\boldsymbol{V}; \Theta)$. From Equation \ref{f(V, Theta)}, we have
\begin{align}
    &\ \ \ \frac{\partial logf(\boldsymbol{V}; \Theta)}{\partial W_{ij}^k} \nonumber \\ 
    &= \frac{1}{f} \frac{\partial f(\boldsymbol{V}; \Theta)}{\partial W_{ij}^k} \nonumber \\
    &= \frac{1}{f} \Sigma_{j'=1}^F \Sigma_{h_{j'}=0}^1 v_i^kh_j exp(\Sigma_{i=1}^m\Sigma_{j=1}^F\Sigma_{k=1}^K v_{i}^k W_{ij}^k h_j \nonumber \\
    &\ \ \ + \Sigma_{i=1}^m\Sigma_{k=1}^K v_i^k b_i^k + \Sigma_{j=1}^F h_j c_j) \nonumber \\
    &= \frac{1}{f} v_i^kh_j f \nonumber \\
    &= v_i^kh_j.
\end{align}
Similarly, it can be shown that
\begin{align}
    &\frac{\partial logf(\boldsymbol{V}; \Theta)}{\partial b_i^k} = v_i^k, \\ 
    &\frac{\partial logf(\boldsymbol{V}; \Theta)}{\partial c_j} = h_j.
\end{align}
Then the gradient descent learning algorithm for RBM is
\begin{align}
    W_{ij}^k \ \leftarrow \ &W_{ij}^k + \eta (\langle v_i^k h_j\rangle_{data} - \langle v_i^k h_j\rangle_{model}), \label{update_W} \\ 
    b_{i}^k \ \leftarrow \ &b_{i}^k + \eta (\langle v_i^k\rangle_{data} - \langle v_i^k\rangle_{model}), \label{update_b} \\
    c_{j} \ \leftarrow \ &c_{j} + \eta (\langle h_j\rangle_{data} - \langle h_j\rangle_{model})) \label{update_c}
\end{align}
where $\eta$ is learning rate and $<\bf{\cdot}>_{data}$, $<\bf{\cdot}>_{model}$ are the expectations corresponding to observed data and the true probability distribution from RBM model respectively.

\subsection{Contrastive Divergence Algorithm with MCMC}
Usually $p(\boldsymbol{V}; \Theta)$ is intractable since $Z(\Theta)$ is unknown. Hence, it is infeasible to analytically compute $\langle v_i^k h_j\rangle_{model}, \langle v_i^k\rangle_{model}$ and $\langle h_j\rangle_{model}$ in the learning algorithm (\ref{update_W}), (\ref{update_b}) and (\ref{update_c}). In practice, Monte Carlo method is applied to compute these expectations approximately, which uses sample mean from a large size sampling set for the joint distribution $p(\boldsymbol{V}, \boldsymbol{H})$ to estimate the theoretical expectations. Since $p(\boldsymbol{V}, \boldsymbol{H})$ is also unknown, it is also infeasible to draw samples from it directly. 

To resolve this difficulty, Hinton et al \cite{005_RBM_Hinton} proposed the contrastive divergence algorithm which utilizes Gibbs sampling technique. It is a MCMC algorithm, to draw samples that asymptotically follow the joint distribution $p(\boldsymbol{V}, \boldsymbol{H})$.



    

\begin{figure}[h]
\centering
\includegraphics[height=2cm, width=9cm]{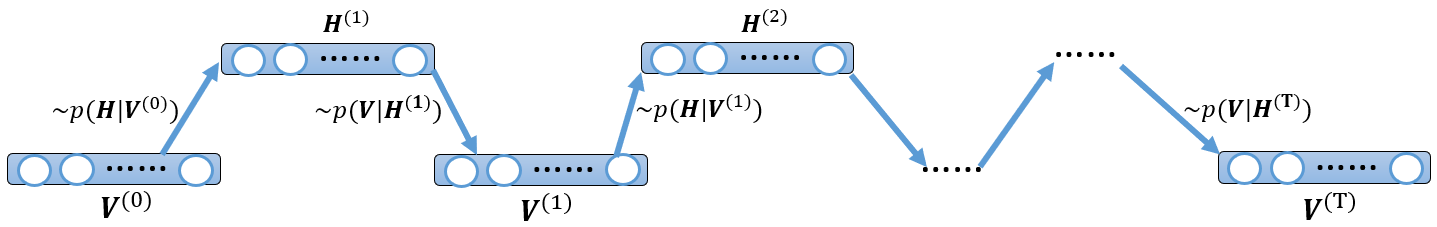}
\caption{Gibbs sampling}
\label{fig:Gibbs_sampling}
\end{figure}

Figure \ref{fig:Gibbs_sampling} illustrates the Gibbs Sampling algorithm. The algorithm generates a Markov chain with known conditional probabilities $p(\boldsymbol{H}|\boldsymbol{V})$ and $p(\boldsymbol{V}|\boldsymbol{H})$. When the chain is long enough, samples at the end of the chain will be theoretically close enough to true samples drawn from the unknown joint distribution $p(\boldsymbol{V}, \boldsymbol{H})$.  In reality, only a few Gibbs steps are enough to generate qualified samples that are needed to estimate the expectations in the learning algorithm (\ref{update_W}), (\ref{update_b}) and (\ref{update_c}). Readers may refer to \cite{MCMC_book} for more details on MCMC and Gibbs sampling.

\subsection{Contrastive Divergence Algorithm for Training RBM}
The contrastive divergence learning algorithm for training RBM via T-step Gibbs sampling is summarized in equations (\ref{Gibbs_W}), (\ref{Gibbs_b}) and (\ref{Gibbs_c}).
\begin{equation}
    W_{ij}^k \ \leftarrow \ W_{ij}^k + \eta (\langle v_i^k h_j\rangle_{data} - \langle v_i^k h_j\rangle_{T-step\ Gibbs\ samples}), \label{Gibbs_W}
\end{equation}
\begin{equation}
    b_{i}^k \ \leftarrow \ b_{i}^k + \eta (\langle v_i^k\rangle_{data} - \langle v_i^k\rangle_{T-step\ Gibbs\ samples}), \label{Gibbs_b}
\end{equation}
\begin{equation}
    c_{j} \ \leftarrow \ c_{j} + \eta (\langle h_j\rangle_{data} - \langle h_j\rangle_{T-step\ Gibbs\ samples}). \label{Gibbs_c}
\end{equation}

\section{Parallelized Training} \label{distributed training}
\begin{figure}[h]
\centering
\includegraphics[scale=0.25]{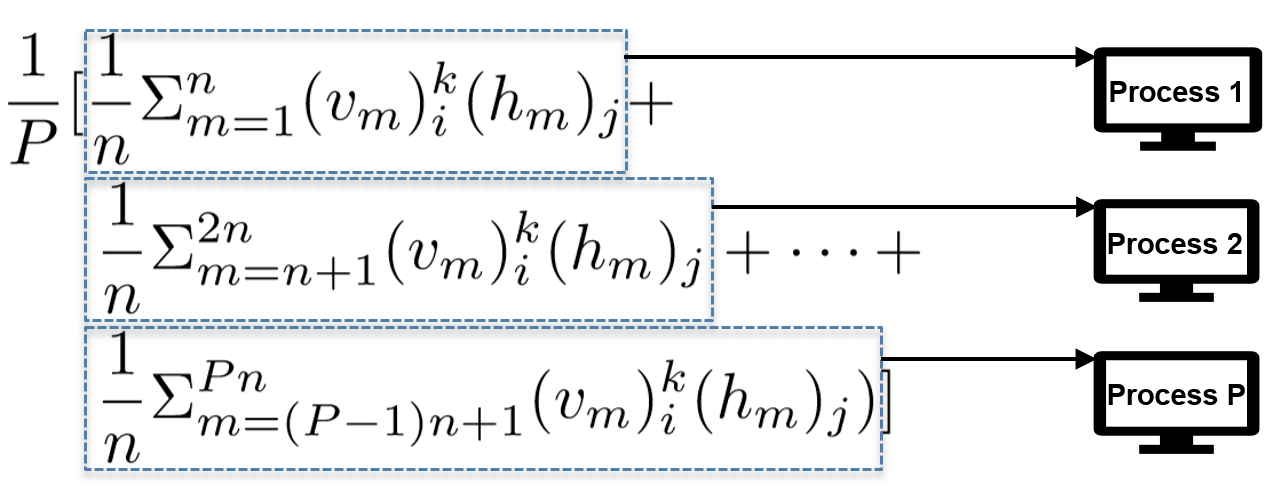}
\caption{Parallel computing of gradients}
\label{fig:parallel}
\end{figure}

The learning algorithms (\ref{Gibbs_W}), (\ref{Gibbs_b}) and (\ref{Gibbs_c}) are parallel by nature. Suppose we have a batch of training data $\{(v_m)_i^k, (h_m)_j\}_{m=1}^N$ where $(v_m)_i^k$ and $(h_m)_j$ are $v_i^k$ and $h_j$ for the $m^{th}$ data point in the training set. If $N$ can be evenly divided into $P$ parts with $N=Pn$, then
\begin{equation}
\begin{aligned}
\langle v_i^k h_j\rangle_{data} &= \frac{1}{N} \Sigma_{m=1}^N (v_m)_i^k (h_m)_j \nonumber \\
                    &= \frac{1}{Pn} \Sigma_{m=1}^{Pn} (v_m)_i^k (h_m)_j \nonumber \\
                    &= \frac{1}{P}[\frac{1}{n} \Sigma_{m=1}^{n} (v_m)_i^k (h_m)_j + \nonumber \\
                                   &\ \ \ \ \ \ \ \ \frac{1}{n} \Sigma_{m=n+1}^{2n} (v_m)_i^k (h_m)_j + \cdots + \nonumber \\
                                   &\ \ \ \ \ \ \ \ \frac{1}{n} \Sigma_{m=(P-1)n+1}^{Pn} (v_m)_i^k (h_m)_j)].
\end{aligned}
\end{equation}
Similarly, it can be shown that other formulas for computing expectations in algorithms (\ref{Gibbs_W}), (\ref{Gibbs_b}) and (\ref{Gibbs_c}) are parallel with respect to training data. So we can distribute the computations in the algorithms over $P$ processes, as is illustrated in Figure \ref{fig:parallel}.

In this paper we have performed distributed training of RBM with the Horovod framework developed by Uber \cite{sergeev2018horovod}. Horovod essentially uses a distibuted optimizer strategy which is basically an optimizer that wraps another tf.Optimizer, using an allreduce operation to average gradient values before applying gradients to model weights. In the method $compute\_gradients$, gradients for different processes are computed via $self.\_optimizer.compute\_gradients(\cdot)$, then averaged via $self.\_allreduce\_grads(\cdot)$. The $compute\_gradients$ method for contrastive divergence algorithm with $T=1$ is shown in Listing \ref{lst:cd_optimizer}. To enable distributed training using Horovod, one simply needs to pass $CDOptimizer$ to $horovod.DistributedOptimizer(\cdot)$ as shown in Listing \ref{lst:hvd_optimizer}. Readers may refer to the Horovods github repository for a thorough guidance for parallel training of neural network models.

  
  
  

\begin{lstlisting}[language=Python, caption={Optimizer for contrastive divergence (T=1)\label{lst:cd_optimizer}}]

class CDOptimizer(base):
    
    def __init__(self):
        pass
    
    # v0 is visible layer states for a batch of training data;
    # v0, v1 are 3d tensors; h0, h1 are 2d tensors (matrices);
    def compute_gradients(self, v0, h0, v1, h1):
                  
        cross1 = tf.einsum('sik,sj->sijk', v0, h0)
        w_pos_grad = tf.reduce_mean(cross1, axis=0)
        
        cross2 = tf.einsum('sik,sj->sijk', v1, h1)
        w_neg_grad = tf.reduce_mean(cross2, axis=0)
                  
        CD = w_pos_grad - w_neg_grad
                  
        g_bv = tf.reduce_mean(v0, axis=0) - tf.reduce_mean(v1, axis=0)
        g_bh = tf.reduce_mean(h0, axis=0) - tf.reduce_mean(h1, axis=0)                  
        
        return [(CD, 'w'), (g_bv, 'bv'), (g_bh, 'bh')]
\end{lstlisting}

\begin{lstlisting}[language=Python, caption={Horovod Distributed Optimizer\label{lst:hvd_optimizer}}]

    opt = CDOptimizer()
    opt = hvd.DistributedOptimizer(opt)
\end{lstlisting}

\section{Inference} \label{inference}
After an RBM model is trained (i.e., $W_{ij}^k, b_i^k$ and $c_{j}$ have been learned from training data), we can make prediction for a user's potential rating for a given item via $p(\boldsymbol{V}; \Theta)$. Let $\boldsymbol{V}^{obs}=\{(v_m)_i^k\}_{i\in \mathcal{I}, k\in \mathcal{K}}$ be the observed ratings for user $m$, where $\mathcal{I}, \mathcal{K}$ are sets for indices $i, k$ for which a rating of $k$ for item $i$ is observed (for example, if a rating $4$ is observed for item $221$, then $4$ is an element in $\mathcal{K}$ and $221$ is an element in $\mathcal{I}$). Given $\boldsymbol{V}^{obs}$, the probability that a user will rate item $i'$ with score $k'$ is:

\begin{equation}
\begin{aligned}
    \ \ &p((v)_{i'}^{k'}=1|\boldsymbol{V}^{obs}) \nonumber \\
    =&\frac{1}{p(\boldsymbol{V}^{obs})} p(\{ (v)_{i'}^{k'} \} \bigcup \boldsymbol{V}^{obs}) \nonumber \\
    \propto&p(\{ (v)_{i'}^{k'} \} \bigcup \boldsymbol{V}^{obs})   \nonumber \\
    =&\frac{1}{Z} \Sigma_{\boldsymbol{H}} exp[\Sigma_{j=1}^{F} \Sigma_{i\in \{i'\}\bigcup\mathcal{I}} \Sigma_{k\in \{k'\}\bigcup\mathcal{K}} v_i^k W_{ij}^k h_j \nonumber \\
    &+ \Sigma_{i\in \{i'\}\bigcup\mathcal{I}} \Sigma_{k\in \{k'\}\bigcup\mathcal{K}} v_i^k b_i^k \nonumber \\
    &+ \Sigma_{j=1}^{F}h_jc_j]   \nonumber \\
    \propto& \Sigma_{\boldsymbol{H}} exp[\Sigma_{i\in \mathcal{I}} \Sigma_{k\in \mathcal{K}} + v_{i'}^{k'} b_{i'}^{k'}]  \nonumber \\
           &\Pi_{j=1}^{F}exp[\Sigma_{i\in \mathcal{I}} \Sigma_{k\in \mathcal{K}} v_i^k W_{ij}^k h_j + v_{i'}^{k'} W_{i'j}^{k'} h_j + h_jc_j] \nonumber \\
    =& exp[v_{i'}^{k'} b_{i'}^{k'}] \nonumber \\
     &\overset{F}{\underset{j=1}{\Pi}} \Sigma_{h_j=0}^1 exp[\Sigma_{i\in \mathcal{I}} \Sigma_{k\in \mathcal{K}} v_i^k W_{ij}^k h_j + v_{i'}^{k'} W_{i'j}^{k'} h_j + h_jc_j]  \nonumber \\
    =& exp[v_{i'}^{k'} b_{i'}^{k'}] \nonumber \\
     &\overset{F}{\underset{j=1}{\Pi}} (1+exp[g(\{v_i^k\}_{i\in \mathcal{I}, k\in \mathcal{K}}; \{W_{ij}^k\})+v_{i'}^{k'} W_{i'j}^{k'}+c_j])\nonumber \\
    =& S(k'; i', \boldsymbol{V}^{obs})
\end{aligned}
\end{equation}
where $g(\{v_i^k\}_{i\in \mathcal{I}, k\in \mathcal{K}}; \{W_{ij}^k\}) = \Sigma_{i\in \mathcal{I}} \Sigma_{k\in \mathcal{K}} v_i^k W_{ij}^k$. Then the predicted rating that user $m$ will give to item $i'$ is the one with the highest $S$ value, i.e.
\begin{equation}
    k_0 = argmax(\{S(k'; i', \boldsymbol{V}^{obs})\}_{k' \in \{1, \cdots, K\}}).
\end{equation}
Readers can refer to \cite{003_RBM_for_CF} for more details about making predictions with RBM model.

\ 
\ 
\section{Experiments} \label{experiments}
\begin{figure}[h]
\centering
\includegraphics[scale=0.35]{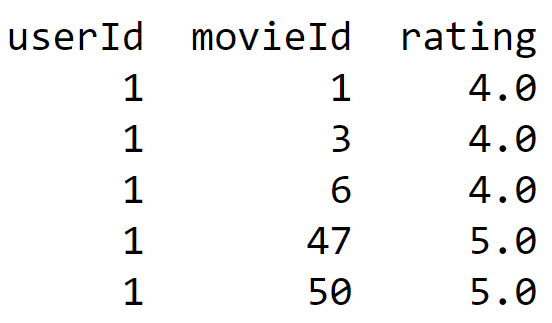}
\caption{MovieLens Data Samples}
\label{fig:data_sample}
\end{figure}

In this section, we test parallelized training of RBM model with the MovieLens data \cite{MovieLens}. The data set has $27,753,444$ ratings for $M=53,889$ movies by $N=283,228$ users. A piece of this data is shown in Figure \ref{fig:data_sample}.

\ 

The rating matrix $\boldsymbol{R}\in \mathbf{R}^{N\times M}$ is defined as
\begin{align}
    \boldsymbol{R}=(r_{nm})_{n\in\{1, \cdots, N\}; m\in \{1,\cdots, M\}}
\end{align}
with $r_{nm}$ being the rating score of user $n$ for movie $m$. If $r_{nm}$ is not observed yet, then we let $r_{nm}=0$ in the rating matrix.


\ 
\subsection{RBM v.s. SVD (Singular Value Decomposition)}
Before testing parallelized training of RBM model, we train RBM model with a small subset of the MovieLens data and compare its performance with that of SVD (Singular Value Decomposition) method, another popular model for collaborative filtering in recommendation system. The small data set has $84,313$ ratings for $9,557$ movies rated by $248$ users. In this test, the first $30$ ratings for each user is held from the data as test set and the remaining part is used as training set.

For SVD method, the basic idea is to keep only a portion of singular values of the rating matrix
\begin{align}
    \boldsymbol{R}=\boldsymbol{U \Lambda V^{T}}
\end{align}
and reconstruct $\boldsymbol{R}$ via 
\begin{align}
    \boldsymbol{R}^{'}=\boldsymbol{U\Lambda^{'}V^{T}}.
\end{align}
Here $\Lambda$ contains all the singular values of $R$ and $\Lambda'$ is obtained by truncating $\Lambda$ and keeping only the first $q$ leading singular values. Then the predicted rating for movie $m'$ by user $n'$ is defined as
\begin{align}
    r^{'}_{n^{'}m^{'}} = \boldsymbol{R}^{'}_{n^{'}m^{'}}.
\end{align}
Readers may refer to \cite{004_SVD_for_CF} for more details on SVD method and its variations.

\begin{figure}[h]
\centering
\includegraphics[scale=0.4]{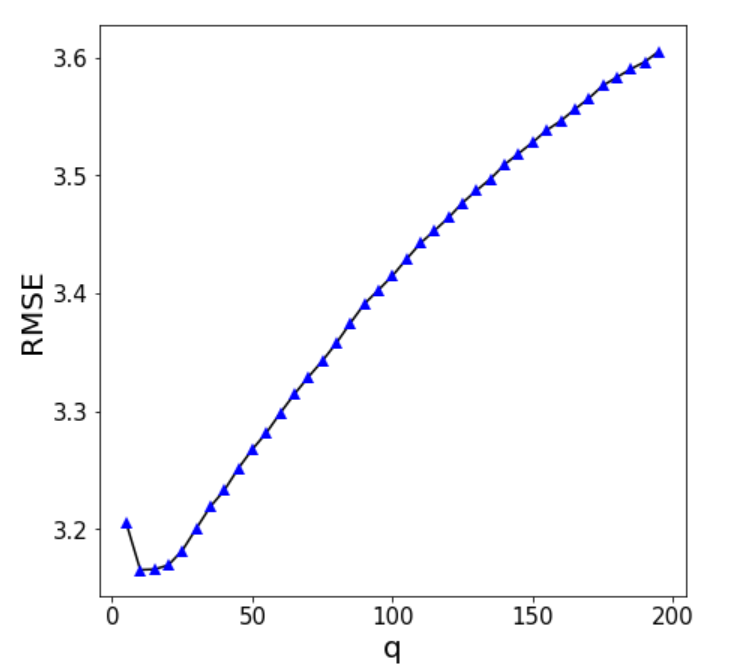}
\caption{$RMSE$ for SVD Models with Different $q$ Values}
\label{fig:SVD}
\end{figure}

Since $q$ is a hyper-parameter for SVD method, we tested SVD models with different values of $q$ and computed their performance metric $RMSE$ (Rooted Mean Square Error). The result is shown in Figure \ref{fig:SVD}. The optimal $q$ value is about $10$.

\begin{figure}[h]
\centering
\includegraphics[scale=0.25]{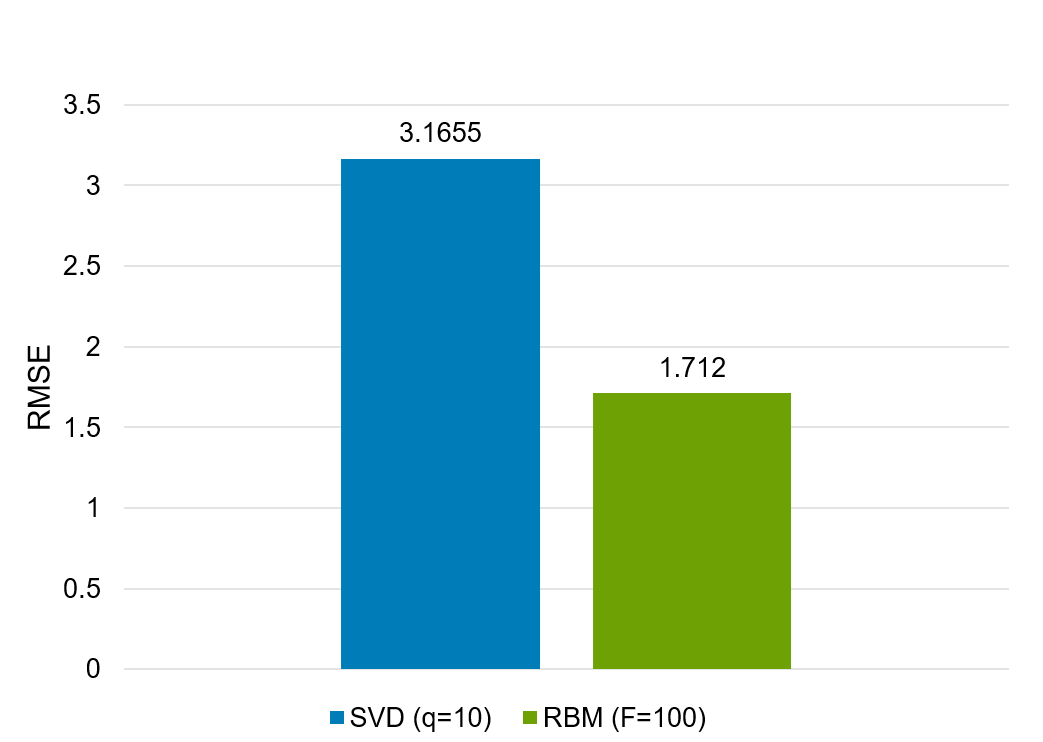}
\caption{Prediction Accuracy Comparison (SVD v.s. RBM)}
\label{fig:SVD_vs_RBM}
\end{figure}

The comparison of prediction accuracy for SVD with $q=10$ and RBM is shown in Figure \ref{fig:SVD_vs_RBM}. The RBM model has $100$ neurons in hidden layer ($F=100$) and is trained with global batch size of $50$ for $100$ epochs with a learning rate of 0.001 on 2 processes in a single compute node.  As is shown in Figure \ref{fig:SVD_vs_RBM}, RMSE value for SVD model with the optimal $q$ value ($q=10$) is still way larger than that for RBM model. In reality, RBM model produces more accurate predictions than SVD model in most cases.

While prediction accuracy of RBM model is usually higher than that of SVD model, training an RBM model could be computationally slow. This motivated us to explore on parallelized training of RBM model.

\ 
\subsection{Parallelized Training of RBM Model}
The data we used to test parallelized training of RBM model is obtained by selecting observed ratings from users who have rated at least $100$ different movies. The selected data set contains $21,595,144$ ratings for $53,324$ movies by $68,342$ users. Similarly, $30$ of the ratings from each user is held as test data. The tests were run on the Zenith supercomputer at Dell EMC HPC \& AI Innovation Lab~\cite{zenith}.

\subsubsection{Strong Scaling}
\begin{figure}[h]
\centering
\includegraphics[width=\columnwidth]{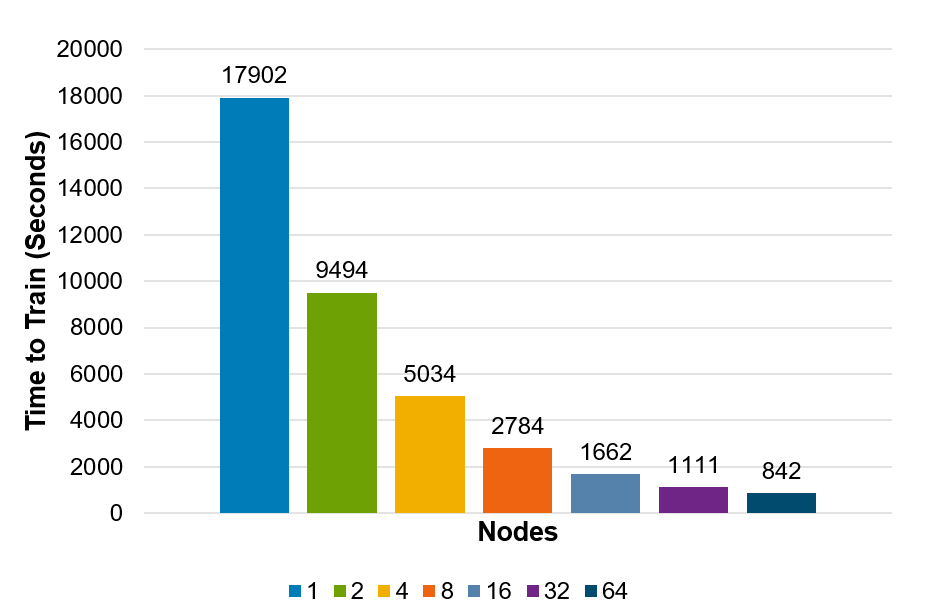}
\caption{Time to Solution (Strong Scaling)}
\label{fig:time_to_train_strong}
\end{figure}

\begin{figure}[h]
\centering
\includegraphics[width=\columnwidth]{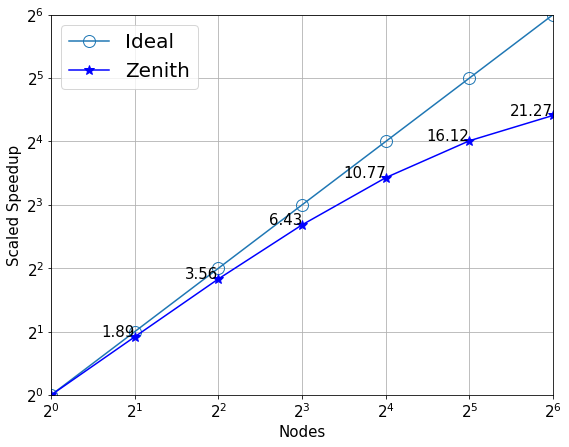}
\caption{Scaled Speedup for Strong Scaling (1 Epoch)}
\label{fig:Scaled_Speedup_Strong_Scaling_1_Epoch}
\end{figure}

To test the performance of strong scaling for RBM model, we fix the global batch size to be $512$. Then we train the model for one epoch with $1, 2, 4, \cdots, 64$ nodes with $1$ process per node. The tests were run for one epoch. Time-to-train and scaled speedup are shown in Figure \ref{fig:time_to_train_strong} and Figure \ref{fig:Scaled_Speedup_Strong_Scaling_1_Epoch} respectively.

\subsubsection{Weak Scaling}
\begin{figure}[h]
\centering
\includegraphics[width=\columnwidth]{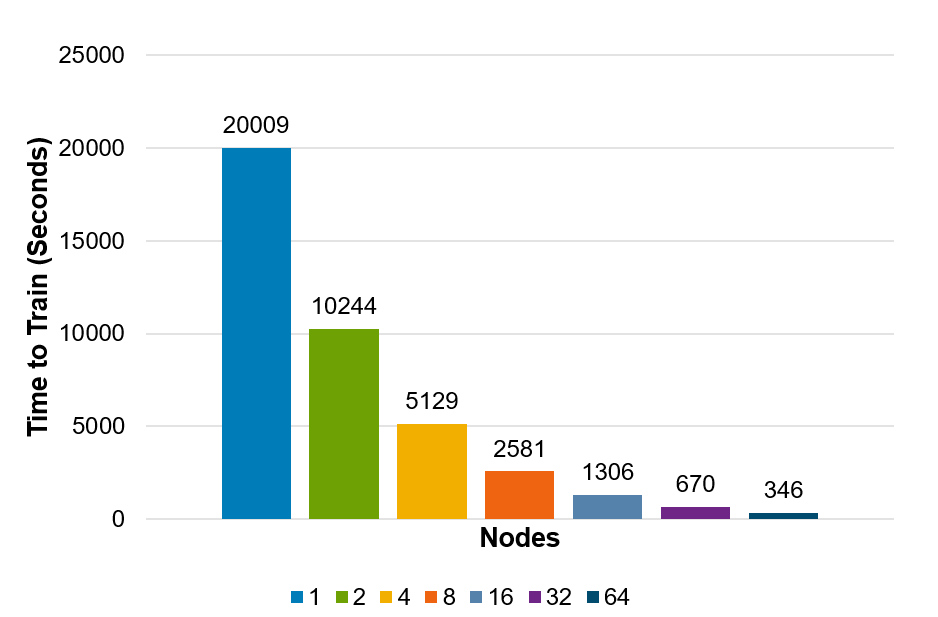}
\caption{Time to Solution (Weak Scaling)}
\label{fig:time_to_train_weak}
\end{figure}

\begin{figure}[h]
\centering
\includegraphics[width=\columnwidth]{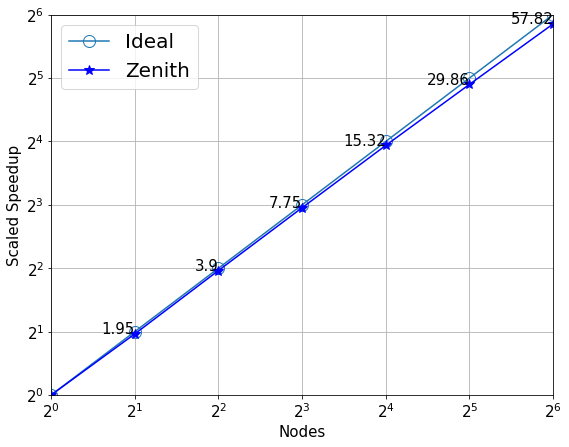}
\caption{Scaled Speedup for Weak Scaling (1 Epoch)}
\label{fig:Scaled_Speedup_Weak_Scaling_10_Epochs}
\end{figure}

For weak scaling test, the batch size for each node is set to be 100. We run the test for one epoch for $1, 2, 4, \cdots, 64$ nodes. Time-to-train and scaled speedup are shown in Figure \ref{fig:time_to_train_weak} and Figure \ref{fig:Scaled_Speedup_Weak_Scaling_10_Epochs} respectively.

\subsubsection{Prediction Accuracy}
We trained an RBM model with global batch size $512$ over $8$ processes for $100$ epochs. The model has $100$ neurons in the hidden layer. The trained model was then applied to make predictions for $10,963$ ratings in the test data set. The $RMSE$ value is about $1.62$. In future work, we will train RBM models with larger global batch size and test the prediction accuracy for them.

\FloatBarrier

\section{Conclusion}\label{conclusion}
In this paper, we studied the principles of RBM model and parallelized training with Horovod framework for it. As is shown in the paper, parallelized training can significantly shorten the training time. Only in this way, RBM models can be practically applied for collaborative filtering in recommendation systems.

Experiments using our technique to train a Restricted Boltzmann Machine with the MovieLens dataset showed that both strong and weak scaling could be maintained out to 64 compute nodes while producing quality models in accordance with the scale of the dataset used. Future work will focus on training at greater scale using larger datasets.

\section{Conflict of Interest}\label{conflict}
On behalf of all authors, the corresponding author states that there is no conflict of interest. 


\bibliographystyle{spmpsci}      

\begin{thebibliography}{10}
\providecommand{\url}[1]{{#1}}
\providecommand{\urlprefix}{URL }
\expandafter\ifx\csname urlstyle\endcsname\relax
  \providecommand{\doi}[1]{DOI~\discretionary{}{}{}#1}\else
  \providecommand{\doi}{DOI~\discretionary{}{}{}\begingroup
  \urlstyle{rm}\Url}\fi

\bibitem{CarreiraPerpin2005OnCD}
Carreira-Perpi{\~n}{\'a}n, M.{\'A}., Hinton, G.E.: On contrastive divergence
  learning.
\newblock In: AISTATS (2005)

\bibitem{zenith}
{Dell EMC}: {HPC \& AI Innovation Lab}.
\newblock
  \url{https://www.dellemc.com/en-us/solutions/high-performance-computing/HPC-AI-Innovation-Lab.htm}
  (2019)

\bibitem{MCMC_book}
Gilks W.~(Ed.), R.S.E.S.D.E.: Markov Chain Monte Carlo in Practice.
\newblock New York: Chapman and Hall/CRC (1996).
\newblock \urlprefix\url{https://doi.org/10.1201/b14835}

\bibitem{MovieLens}
Harper, F.M., Konstan, J.A.: Acm transactions on interactive intelligent
  systems (tiis) 5, 4, article 19 (december 2015), 19 pages.
\newblock The MovieLens Datasets: History and Context  (2015)

\bibitem{005_RBM_Hinton}
Hinton, G.: https://www.cs.toronto.edu/~hinton/csc321/readings/boltz321.pdf
  (2017)

\bibitem{006_RBM_Learning_ReLearning}
Hinton, G.E., Sejnowski, T.J.: Learning and relearning in boltzmann machines
  (1986)

\bibitem{003_RBM_for_CF}
Ruslan~Salakhutdinov Andriy~Mnih, G.H.: Restricted boltzmann machine for
  collaborative filtering.
\newblock Proceedings of the 24 th International Conference on Machine Learning
   (2007)

\bibitem{002_Deep_Boltzmman_Machine}
Salakhutdinov, R., Hinton, G.: Deep boltzmann machines.
\newblock In: D.~van Dyk, M.~Welling (eds.) Proceedings of the Twelth
  International Conference on Artificial Intelligence and Statistics,
  \emph{Proceedings of Machine Learning Research}, vol.~5, pp. 448--455. PMLR,
  Hilton Clearwater Beach Resort, Clearwater Beach, Florida USA (2009).
\newblock \urlprefix\url{http://proceedings.mlr.press/v5/salakhutdinov09a.html}

\bibitem{sergeev2018horovod}
Sergeev, A., Balso, M.D.: Horovod: fast and easy distributed deep learning in
  {TensorFlow}.
\newblock arXiv preprint arXiv:1802.05799  (2018)

\bibitem{001_Smolensky}
Smolensky, P.: Parallel distributed processing: Explorations in the
  microstructure of cognition, vol. 1.
\newblock chap. Information Processing in Dynamical Systems: Foundations of
  Harmony Theory, pp. 194--281. MIT Press, Cambridge, MA, USA (1986).
\newblock \urlprefix\url{http://dl.acm.org/citation.cfm?id=104279.104290}

\bibitem{004_SVD_for_CF}
Xian, Z., Li, Q., Li, G., Li, L.: New collaborative filtering algorithms based
  on svd++ and differential privacy.
\newblock Mathematical Problems in Engineering \textbf{2017}, 1--14 (2017).
\newblock \doi{10.1155/2017/1975719}

\bibitem{004_Yu}
Yu, H.: A gentle tutorial on restricted boltzmann machine and contrastive
  divergence  (2017).
\newblock \doi{10.13140/RG.2.2.26119.60326}

\end{thebibliography}

\end{document}